\documentclass[10pt,twocolumn,letterpaper]{article}

\usepackage[pagenumbers]{cvpr} %

\usepackage{graphicx}
\usepackage{multirow}
\usepackage{makecell}
\usepackage{amssymb}
\usepackage{xcolor}
\usepackage{algorithm}
\usepackage{algorithmic}
\usepackage{duckuments}
\usepackage{colortbl}

\def\etal{\emph{et al.}}

\definecolor{cvprblue}{rgb}{0.21,0.49,0.74}

\definecolor{Gray}{gray}{0.85}
\definecolor{brown}{rgb}{0.65, 0.16, 0.16}
\definecolor{purple}{rgb}{0.65, 0.1, 0.75}
\definecolor{yellow}{rgb}{0.75, 0.75, 0.0}
\definecolor{orange}{rgb}{1.0, 0.5, 0.2}
\definecolor{green}{rgb}{0, 0.8, 0.2}
\definecolor{red}{rgb}{1.0, 0.0, 0.0}
\definecolor{dblue}{rgb}{0.6, 0.1, 0.9}
\definecolor{black}{rgb}{0.0, 0.0, 0.0}

\usepackage[pagebackref,breaklinks,colorlinks,allcolors=cvprblue]{hyperref}

\def\paperID{6918} %
\def\confName{CVPR}
\def\confYear{2025}

\title{Improving Sound Source Localization with\\Joint Slot Attention on Image and Audio}

\author{Inho Kim$^{1}$ \quad
Youngkil Song$^{2}$ \quad
Jicheol Park$^{2}$ \quad
Won Hwa Kim$^{1, 2}$ \quad
Suha Kwak$^{1, 2}$ \vspace{1mm} \\
$^{1}$Dept. of CSE, POSTECH \qquad $^{2}$Graduate School of AI, POSTECH \\
{\tt\{\small{kimih, songyk, jicheol, wonhwa, suha.kwak}\}@postech.ac.kr}
}

\begin{document}
\maketitle
\begin{abstract}
Sound source localization (SSL) is the task of locating the source of sound within an image. 
Due to the lack of localization labels, the de facto standard in SSL has been to represent an image and audio as a single embedding vector each, and use them to learn SSL via contrastive learning. 
To this end, previous work samples one of local image features as the image embedding and aggregates all local audio features to obtain the audio embedding, which is far from optimal due to the presence of noise and background irrelevant to the actual target in the input.
We present a novel SSL method that addresses this chronic issue by joint slot attention on image and audio.
To be specific, two slots competitively attend image and audio features to decompose them into target and off-target representations, and only target representations of image and audio are used for contrastive learning.
Also, we introduce cross-modal attention matching to further align local features of image and audio.
Our method achieved the best in almost all settings on three public benchmarks for SSL, and substantially outperformed all the prior work in cross-modal retrieval.
\end{abstract}
    
\section{Introduction}
\label{sec:intro}
Humans easily identify where a sound comes from within their sight.
This capability is a key aspect of human perception facilitating interaction with  surroundings, and is also demanded in a large variety of vision applications such as autonomous driving and robotics~\cite{furletov2021auditory, sasaki2018online, yin2023real}.
Motivated by this, a large body of research~\cite{objects_eccv18, attention_cvpr18, dmc_cvpr19, coarsetofine_eccv20, avobject_eccv20, lvs_cvrp21, ezvsl_eccv22, hardpos_icassp22, slavc_neurips22, sspl_cvpr22, fnac_cvpr23, alignment_iccv23, liu_fullsup_acmmm23, clip_wacv24, noprior_cvpr24} has studied sound source localization (SSL), the task of locating the source of a sound within the paired image in the form of segmentation mask.

\begin{figure}[t]
    \centering
    \includegraphics[width=0.942\columnwidth]{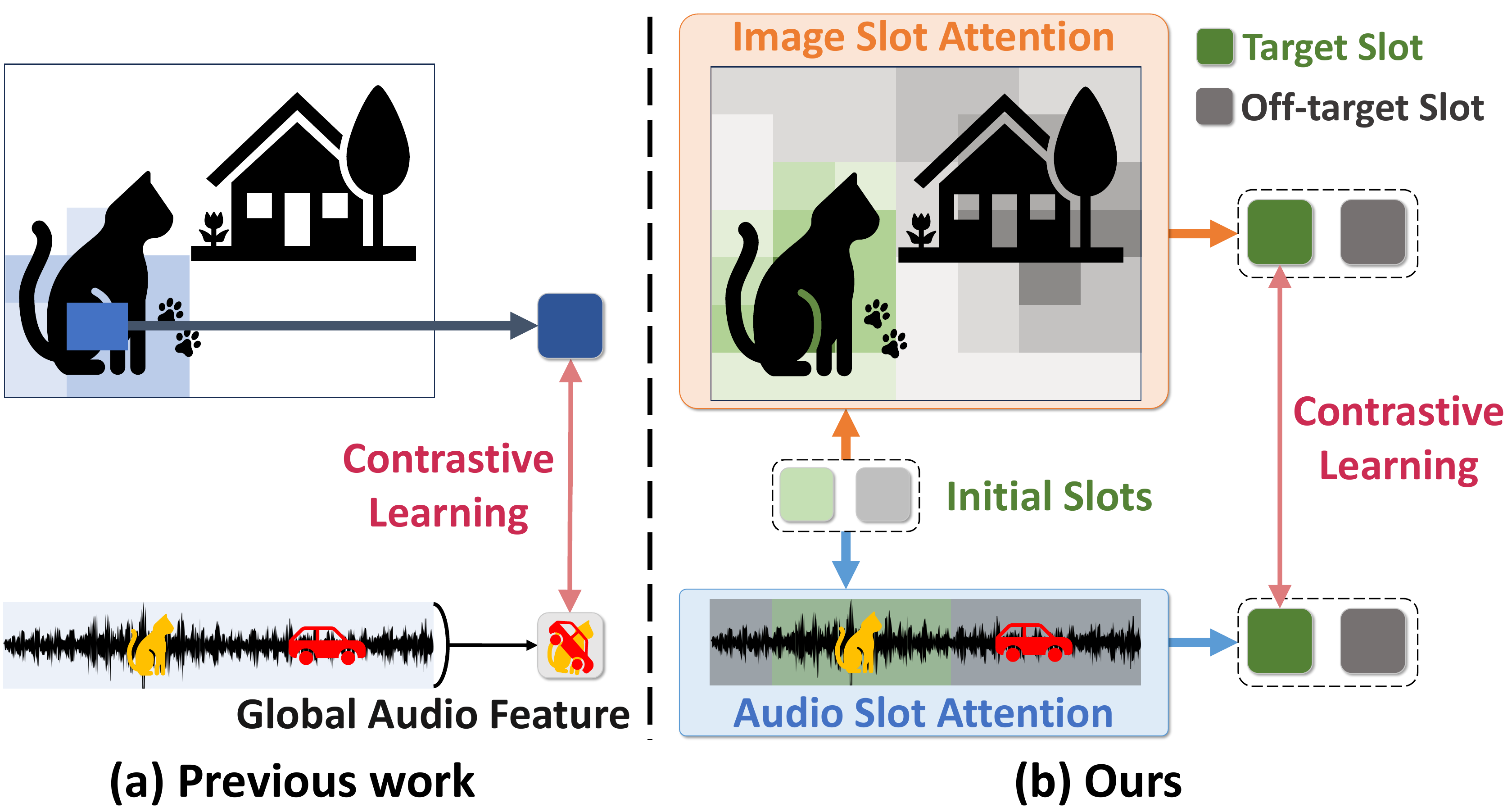}
    \caption{\textbf{Difference between previous work and ours.} 
    For contrastive learning of SSL, our method identifies and utilizes pairs of relevant image region and audio segment, while previous work exploits a local image feature and the global audio feature.
    }
    \vspace{-1mm}
    \label{concept_fig}
\end{figure}

Due to the prohibitively expensive cost of sound source annotation for image-audio pair data, existing methods~\cite{alignment_iccv23,ezvsl_eccv22,lvs_cvrp21,fnac_cvpr23} have focused on learning SSL using only image-audio pairs without sound source labels.
To this end, they consider the task as multiple instance learning~\cite{maron1997framework, ilse2018attention, andrews2002support} and assume that, given an image-audio pair, at least one patch of the image is relevant to the audio; then they represent the image with its local feature that most closely resembles the global audio feature, and learn SSL through contrastive learning~\cite{nce_arxiv18, clip_pmlr21} with these features (Fig.~\ref{concept_fig}(a)).

While this approach has been proven to work, there is plenty of room for further improvement.
First, audio may contain not only the sound of the target shared by the two modalities but also irrelevant sounds, such as noise or sounds from sources outside the image.
In such cases, using the global audio feature for contrastive learning 
hinders accurate alignment between the image and audio features.
Second, representing an image with only a single local feature in contrastive learning causes models to locate a small discriminative part of the sound source, not its whole body~\cite{Cam}, which can degrade SSL accuracy. %

\begin{figure*}[t]
    \centering
    \includegraphics[width=0.97\textwidth]{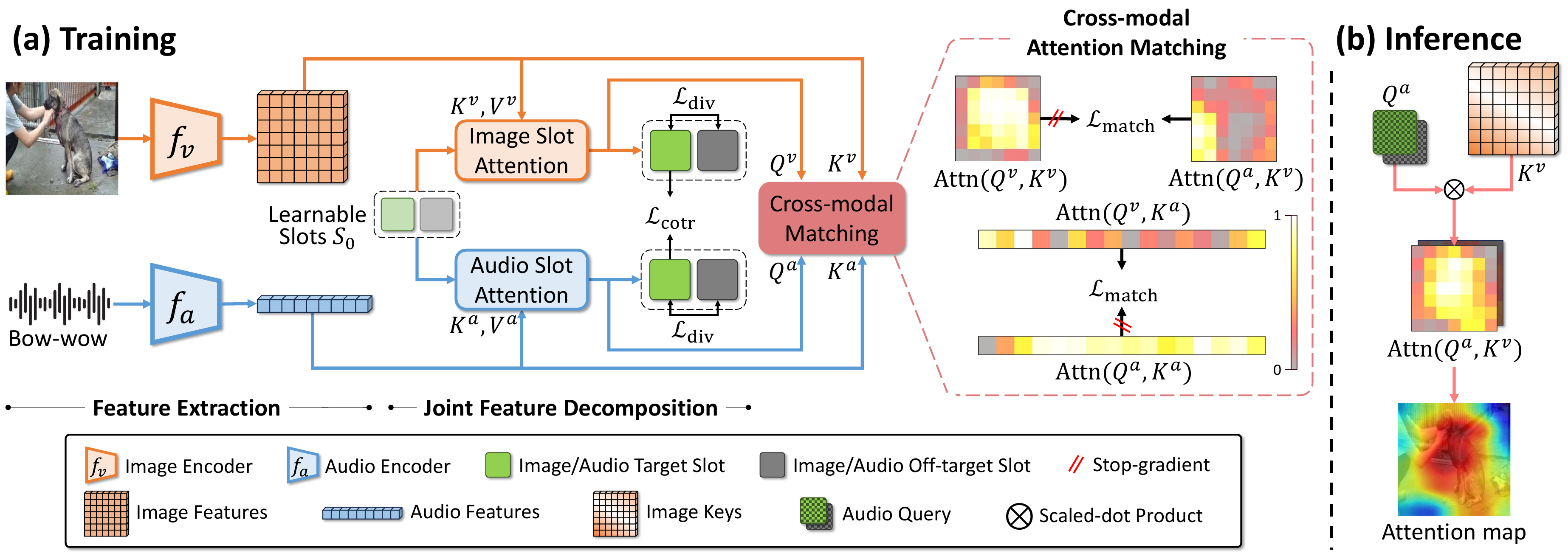}
    \caption{\textbf{Overview of the proposed method.} 
    First, image and audio features are extracted by the image and audio encoders, and serve as keys and values for slot attention. In slot attention, two learnable slots are used as queries to draw attentions on the input features while competing with each other, decomposing the features into the target and off-target representations per modality.
    The model is then trained by contrastive learning using only the target representations, and by cross-modal attention matching for further feature-level alignment between the two modalities.
    For localization in testing, we draw the attention map between the audio target representation and image features.
    Some details for learning slot attention, such as decoders and loss for slot reconstruction, have been omitted for clarity.
    }
    \label{main_fig}
\end{figure*}

To address these limitations, we propose a novel SSL method based on slot attention~\cite{slotattn_neurips20}, which is conceptually illustrated in Fig.~\ref{concept_fig}(b).
In slot attention, slots, which are randomly sampled or learnable embeddings, compete with each other for aggregating input features progressively, and thus they decompose the features into distinct and exclusive representations. 
Motivated by this, we employ
slot attention with two learnable slots to jointly decompose image and audio features into two distinct slot representations per modality, 
one for the target and the other for the rest.
To be specific, we designate the first slot as the \textit{target slot}, which is guided to focus on information related to the shared sound source across both modalities.
The exclusive assignment property of slot attention directs the second slot, referred to as the \textit{off-target slot}, to capture remaining elements, such as the image background or irrelevant sounds in the audio.
Then we learn SSL by contrastive learning using the two target slot representations from the image and audio, respectively. 
Unlike the prior work using only a single global audio feature, our method allows to exploit only the relevant part (\ie, the target slot) of the audio for more accurate SSL.
Furthermore, since the target slot of the image covers the sound source area rather than a single location, our method avoids the problem of localization being reduced to a small discriminative part.

Given the target slot representations, SSL is conducted by drawing an attention map on the image feature map using the target slot representation of the audio as query (Fig.~\ref{main_fig}(b)).
For accurate pixel-level localization, the target slot of the audio should be well aligned with the keys derived from relevant individual image features; in this regard, contrastive learning using only the target slots is insufficient as it does not relate the audio target slot with individual image features.
We thus introduce an additional training strategy dubbed {\it cross-modal attention matching}.
This strategy encourages the cross-modal attention map between the audio target slot and the input image features to approximate the intra-modal attention map between the image target slot and the input image features that are inherently well aligned.
In other words, it provides pseudo-supervision for SSL prediction (\ie, the cross-modal attention map) using the intra-modal attention map, which is drawn by the query and keys from the same modality and thus better localizes the target.
To further facilitate its cross-modal alignment, we apply this strategy to the audio modality as well.

Our model was trained on two datasets, Flickr-SoundNet~\cite{soundnet_neurips16} and VGG-Sound~\cite{vggsound_icassp20}, and evaluated on three benchmarks, Flickr-SoundNet-Test~\cite{soundnet_neurips16}, VGG-SS~\cite{vggsound_icassp20}, and AVSBench~\cite{avsbench_eccv22}. 
It achieved the state of the art in almost all settings across all the three evaluation benchmarks.
Further, our model outperformed all the prior work by large margin in cross-modal retrieval, suggesting that it is not a simple saliency detector but properly learns to relate image and audio for SSL~\cite{alignment_iccv23}.
In summary, the contribution of our method is three-fold:

\begin{itemize}
    \item We propose a novel SSL model that identifies and extracts features relevant to sound source jointly from image and audio modalities through slot attention.
    \item We also propose cross-modal attention matching, which enables effective cross-modal alignment for individual input features and thus improves localization performance.
    \item Our model achieved the state of the art in almost all settings across all the three SSL benchmarks and improved cross-modal retrieval performance substantially.
\end{itemize}

\section{Related Work} \label{sec:related}
Sound source localization aims to identify specific regions in an image that correspond to the sources of sounds from an audio input.
Since constructing audio-visual datasets is costly, sound source localization studies have evolved into self-supervised learning approaches that use only image-audio pairs for contrastive learning~\cite{attention_cvpr18, dmc_cvpr19, coarsetofine_eccv20, avobject_eccv20, lvs_cvrp21,ezvsl_eccv22, hardpos_icassp22, slavc_neurips22, ssltie_acmmm22, fnac_cvpr23, alignment_iccv23, noprior_cvpr24}.

An early study, ~\cite{attention_cvpr18}, addresses the task through contrastive learning between the audio-aware image features and the global audio feature.
The audio-aware image features are obtained by calculating a correspondence map through the normalized inner product between the global audio feature and the image feature map.
Chen \etal~\cite{lvs_cvrp21} propose method that uses a differentiable threshold on the correspondence map to divide audio-aware and other image features, treating the latter as hard negatives in contrastive learning to improve localization.

Recent studies focus on fine-grained sound source localization through contrastive learning between the global audio feature and the most relevant feature selected from the image feature map ~\cite{ezvsl_eccv22, slavc_neurips22, fnac_cvpr23, alignment_iccv23}.
Mo and Morgado~\cite{ezvsl_eccv22} propose a method that refines the audio-visual correspondence map through a post-processing technique leveraging a pretrained visual model.
Sun \etal~\cite{fnac_cvpr23} introduce a regularizer that applies mean absolute error regularization to inter-modal and intra-modality similarity matrices to increase awareness of false negatives.
Senocak \etal~\cite{alignment_iccv23} propose a cross-modal alignment method that preserves localization performance and enhances cross-modal retrieval by leveraging multi-view augmentation and nearest neighbor search to construct multiple positive pairs for contrastive learning.
These methods employ contrastive learning with the most relevant image feature to assist in coarsely identifying sound sources but often fall short of capturing a comprehensive representation, thereby hindering localization ability.
Furthermore, relying solely on global audio features can mislead contrastive learning when noise exists, potentially highlighting incorrect image features.

Following a complementary direction, several studies employ external prior knowledge or even utilize off-the-shelf segmentation modules to achieve high-quality localization results~\cite{coarsetofine_eccv20, clip_wacv24}.
Also, several studies employ sound source localization with additional information such as optical flow~\cite{heartheflow_wacv23}, text inputs~\cite{tvsl_cvpr24} or ground truth category~\cite{avgn_cvpr23}.
By contrast to these studies, our method can effectively perform sound source localization without relying on external prior knowledge or off-the-shelf models.

To address the above issues, we propose a sound source localization framework based on the slot attention mechanism.
Slot attention~\cite{slotattn_neurips20} is proposed for unsupervised object-centric learning, effectively decomposing input features into distinct and mutually exclusive components via iterative attention and competition among slots.
Inspired by this, we adopt slot attention in both image and audio features to decompose them into features related to the sound source and those irrelevant to it without supervision.

\section{Proposed Method}
As shown in Fig.~\ref{main_fig}, our model consists of two parts: 1) image and audio encoders for feature extraction, and 2) joint feature decomposition module with joint slot attention mechanism.
A pair of image and audio is first fed into each encoder to extract image and audio features (Sec.~\ref{method:ext}).
The features of each modality are decomposed into two distinct slot representations by using joint slot attention: one slot for the target and the other for the rest. (Sec.~\ref{method:ssdm}).
The slot representations are then used for training as input to contrastive learning and cross-modal attention matching (Sec.~\ref{method:learning}),
while effectively handling false negatives (Sec.~\ref{method:migitate}).
Lastly, an attention on the image feature map using the target slot representation of the audio is computed to locate the source of sound (Sec.~\ref{method:infer}).

\label{sec:method}
\subsection{Feature Extraction} \label{method:ext}
Let $D = \{(v_i, a_i)\}_{i=1}^{M}$ denote a training set of image-audio pairs, where $v_i \in \mathbb{R}^{3 \times H_v \times W_v}$ is an image of size $H_v \times W_v$, $a_i \in \mathbb{R}^{1 \times F_a \times T_a}$ is a corresponding audio spectrogram of frequency resolution $F_a$ and temporal length $T_a$, and $M$ denotes the total number of such pairs in the dataset. For feature extraction, separate image and audio encoders $f_v$ and $f_a$, are adopted.
The image encoder extracts the image features  $ \mathbf{v}_i := f_v(v_i) \in \mathbb{R}^{h \times w \times c}$ for $v_i$, and the audio encoder extracts the audio features $\mathbf{a}_i := f_a(a_i) \in \mathbb{R}^{t \times c}$ for $a_i$, where $h$ and $w$ are the height and width of the image feature map respectively, $t$ is the length of the audio features, and $c$ is the image and audio feature dimension.

\subsection{Joint Feature Decomposition} \label{method:ssdm}
Slot attention~\cite{slotattn_neurips20} is an attention mechanism where each slot progressively competes to aggregate input features, leading to the decomposition of input features into distinct and exclusive representations.
Building on this idea, for joint feature decomposition, we propose \textit{joint slot attention (JSA)}, which jointly decomposes image and audio features into two distinct slot representations for each modality.
To achieve this, we employ slot attention with two initial slots $S_0 \in \mathbb{R}^{2 \times D}$ shared in the two modalities. %
Furthermore, we designate the first slot as \textit{target slot}, which captures features relevant to sound source, and the other as \textit{off-target slot}, which captures features for the rest.
Since the overall pipeline of JSA is identical in the two modalities, we will describe the module in terms of image only for brevity. 

JSA first projects an input image feature $\mathbf{v}$ to derive its corresponding key $K^v$ and value $V^v$ by applying the following operations: 
\begin{equation}
\begin{split}
    K^v = W^k(\text{LayerNorm}(\mathbf{v})) \in \mathbb{R}^{hw \times d},\\
    V^v = W^v(\text{LayerNorm}(\mathbf{v})) \in \mathbb{R}^{hw \times d},\\
\end{split}
\end{equation}
where $W^k$, $W^v$ are the linear projection layers to obtain embeddings in $d$-dimension.
For each iteration $n \in [1, N]$, the query $Q^v_n$ is derived by applying linear projection to the slots with the initial slots $S_0^v := S_0$, which is given by
\begin{equation}
    Q_n^v = W^q(\text{LayerNorm}(S_{n-1}^v)) \in \mathbb{R}^{2 \times d},
\end{equation}
where $W^q$ is a linear projection layer shared for all iterations.
The slot attention matrix $A$ is computed by taking the dot product of the given query $Q^v_n$ and key $K^v$, scaled by the square root of the dimension $d$
as follows:
\begin{equation}
    A_{i, j} = \dfrac{\text{exp}(M_{i, j})}{\sum_{l=1}^{2}\text{exp}(M_{i, l})}, ~\text{where}~ M = \dfrac{K^v (Q_n^v)^{\top}}{\sqrt{d}} \in \mathbb{R}^{hw \times 2},
\label{eq:attn}
\end{equation}
where $i$ is key index, and $j$ is slot index. Subsequently, a weighted mean is utilized to the slot attention matrix $A$:
\begin{equation} \hat{A}_{i, j} = \frac{A_{i, j}}{\sum_{l=1}^{hw}A_{l, j}}.
\label{eq:slot_norm}
\end{equation}
Then, the updated slot $\bar{S_n}$ is computed by Gated Recurrent Unit (GRU)~\cite{gru_emnlp14} with $S_{n-1}$ as hidden state, and $\hat{A}^{\top} \cdot V^v$ as input as follows:
\begin{equation}
    \bar{S_n^v} = \text{GRU}(S_{n-1}^v,~\hat{A}^{\top}\cdot V^v).
\end{equation}
Finally, the slot is updated by a residual connection:
\begin{equation}
    S_n^v = \bar{S_n^v} + \text{MLP}(\text{LayerNorm}(\bar{S_n^v})) \in \mathbb{R}^{2 \times d}.
\end{equation}
The overall process encourages the input feature to bind to specific slots, ensuring that each slot forms a mutually exclusive representation. The final outputs of JSA are image slots $\{\mathbf{p}^v;\mathbf{r}^v\} := S^v_N \in \mathbb{R}^{2 \times d}$ and image query $Q_N^v \in \mathbb{R}^{2 \times d}$, where $\mathbf{p}^v$ is a target slot and $\mathbf{r}^v$ is an off-target slot.
The same process is applied to the audio modality using audio feature $\textbf{a}$, yielding audio slots $\{\mathbf{p}^a;\mathbf{r}^a\} := S^a_N \in \mathbb{R}^{2 \times d}$ and audio query $Q_N^a \in \mathbb{R}^{2 \times d}$.
Since this process begins from the shared initial slots, the initial slots are able to effectively learn joint features between two modalities.
For brevity, we omit the subscript $N$ for the final query, $Q^v := Q_N^v$ and $Q^a := Q_N^a$.

\subsection{Learning Objectives} \label{method:learning}
\noindent \textbf{Contrastive Learning with Target Slots.} 
We adopt the InfoNCE loss~\cite{nce_arxiv18, clip_pmlr21} for contrastive learning using the target slots, guiding the target slots to identify features relevant to the sound source jointly from image and audio. Let $s$ denote a similarity function, $s(x, y) = \text{exp}(\text{cos}(x, y) / \tau)$, where $\tau$ is a temperature hyperparameter.
The contrastive learning loss $\mathcal{L}_{\text{cotr}}$ is given by
\begin{equation}
    \mathcal{L}_{\text{cotr}} = -\dfrac{1}{B}\sum_{i=1}^{B}\left\{\text{log}\dfrac{s(\mathbf{p}_i^v, \mathbf{p}_i^a)}{\sum_{j}^B s(\mathbf{p}_i^v, \mathbf{p}_j^a)} + \text{log}\dfrac{s(\mathbf{p}_i^v, \mathbf{p}_i^a)}{\sum_{j}^B s(\mathbf{p}_j^v, \mathbf{p}_i^a)}\right\},
\label{eq:cotr}
\end{equation}
where $B$ denotes the number of data in a mini-batch, and the negative pairs refer to the combinations of visual and audio features that do not belong to the same original pair within the mini-batch.
Since the image target slot captures the whole features of the target region, it avoids focusing solely on a specific discriminative part of the image. Furthermore, the audio target slot captures only the relevant sound from the source, it prevents the issue of learning sounds that are unrelated to the source.

\noindent \textbf{Cross-modal Attention Matching.}
As will be described in Sec.~\ref{method:infer}, the localization result is derived by utilizing cross-modal attention between the audio target slot and image features.
For accurate sound source localization, the audio target slot should be properly aligned with keys computed from the relevant image features.
Although contrastive learning encourages alignment of target slot representations between different modalities, it does not ensure alignment between the audio target slot and image features, which degrades the localization accuracy.
We resolve this issue by \textit{cross-modal attention matching}, which encourages the cross-modal attention map to resemble the intra-modal attention map for improving feature-level alignment across the two modalities.
To this end, we first calculate the cross-modal and intra-modal attention matrices.
Specifically, we compute the cross-modal attention matrix between the audio and visual modalities in the same manner as Eq.~\eqref{eq:attn} and Eq.~\eqref{eq:slot_norm}, with the audio queries \( Q^a \) and visual keys \( K^v \).
We denote this cross-modal attention matrix as \( \hat{A}^{a,v} \in \mathbb{R}^{hw \times 2} \), where \( hw \) is the length of \( K^v \).
Subsequently, to obtain the \textit{cross-modal attention} that represents the attention weights between the audio target query and visual keys, we utilize only the first column of \( \hat{A}^{a,v} \). This cross-modal attention is defined as follows:

\vspace{-2mm}
\[
\textbf{ca}^{a,v} := [\hat{A}^{a,v}_{i,1}]_{i=1}^{hw} \in \mathbb{R}^{hw}.
\]
Similarly, for intra-modal attention matrix within the visual modality, we compute the intra-attention matrix \( \hat{A}^{v,v} \) using the \( Q^v \) and  \( K^v \).
Then, the \textit{intra-modal attention} that represents the attention weights between the image target query and visual keys is defined as follows:
\[
\textbf{ia}^{v,v} := [\hat{A}^{v,v}_{i,1}]_{i=1}^{hw} \in \mathbb{R}^{hw}.
\]
To further facilitate cross-modal alignment, we apply this strategy symmetrically to the audio modality as well.
Finally we formulate the cross-modal attention matching loss as follows:
\begin{equation}
\mathcal{L}_{\text{match}} = \|\textbf{ca}^{a,v} - \text{sg}(\textbf{ia}^{v,v})\|_2^2 + \|\textbf{ca}^{v,a} - \text{sg}(\textbf{ia}^{a,a})\|_2^2,
\end{equation}
where sg(\(\cdot\)) denotes the stop-gradient operation.
$\mathcal{L}_{\text{match}}$ minimizes the discrepancy between cross-modal and intra-modal attentions, promoting robust feature-level alignment between the two modalities for accurate localization.

\noindent \textbf{Slot Divergence.} 
In slot attention, the diversity between the target slot and off-target slot should be large enough to ensure that each slot effectively captures distinct, mutually exclusive regions.
Therefore, we introduce a slot divergence loss that encourages greater diversity among the slots.
The slot divergence loss $\mathcal{L}_{\text{div}}$ is given by 
\begin{equation}
    \mathcal{L}_{\text{div}} = \text{max}\{0, \text{cos}(\mathbf{p}^v, \mathbf{r}^v)\} + \text{max}\{0, \text{cos}(\mathbf{p}^a, \mathbf{r}^a)\}.
\label{eq:div}
\end{equation}
It ensures that the different slot embeddings to be sufficiently dissimilar, enabling the target slot to more effectively localize the sound source.

\noindent \textbf{Slot Reconstruction.}
The successful feature reconstruction in slot attention implies that each slot successfully captures unique and independent parts of the input data, which means that the slot attention mechanism effectively decomposed the input into individual slot representations.
It indicates that each slot accurately represents a specific object or region in the input.
To encourage this condition, we apply the slot reconstruction loss $\mathcal{L}_{\text{recon}}$ as follows:
\begin{equation}
    \mathcal{L}_{\text{recon}} = \|\textbf{v} - g_v(\{\mathbf{p}^v;\mathbf{r}^v\})\|_2^2 + \|\textbf{a} - g_a(\{\mathbf{p}^a;\mathbf{r}^a\})\|_2^2,
\label{eq:recon}
\end{equation}
where $g_v$ and $g_a$ are the decoders for the image and audio features, respectively.
These decoders are composed of a small number of MLP layers with ReLU activations.
Note that the decoders are auxiliary modules used only in training and not required for inference.

The overall loss is a combination of $\mathcal{L}_{\text{cotr}}$, $\mathcal{L}_{\text{match}}$, $\mathcal{L}_{\text{div}}$, and  $\mathcal{L}_{\text{recon}}$, which is given by
\begin{equation}
    \mathcal{L}_{\text{total}} = \mathcal{L}_{\text{cotr}} + \lambda_{1} \mathcal{L}_{\text{match}} + \lambda_{2} \mathcal{L}_{\text{div}} + \lambda_{3} \mathcal{L}_{\text{recon}},
\label{eq:total}
\end{equation}
where $\lambda_{1}$, $\lambda_{2}$, and $\lambda_{3}$ are hyperparameters weighting the loss terms.

\subsection{Mitigating False Negatives} \label{method:migitate}
In general, unpaired images and audios are all treated as negatives for contrastive training.
However, some of them are false negatives that are semantically well matched, so treating these as negatives causes adverse effect~\cite{fnac_cvpr23}.
Therefore, we propose a false negative mitigation strategy using $k$-reciprocal nearest neighbors of target slots.

\renewcommand{\arraystretch}{0.95}
\begin{table*}[!t]
\centering
\begin{minipage}{0.49\linewidth}
    \centering
    \resizebox{\columnwidth}{!}{%
    \begin{tabular}{cc|c|cccc}
    \toprule
    \multirow{2}{*}{\textbf{Trainset}}    & \multirow{2}{*}{\textbf{Method}} & \multirow{2}{*}{OGL} & \multicolumn{2}{c|}{\textbf{VGG-SS}}     & \multicolumn{2}{c}{\textbf{Flickr-SoudNet}} \\
                                 & & & cIoU & \multicolumn{1}{c|}{AUC} & cIoU         & AUC         \\ \hline\hline
    \multirow{10}{*}{\makecell{Flickr\\SoundNet\\10k}}  & Attention10k~\cite{attention_cvpr18} & & -     & -     & 43.60 & 44.90 \\
                                 & CoursetoFine~\cite{coarsetofine_eccv20} & & -     & -     & 52.20 & 49.60 \\
                                 & AVObject~\cite{avobject_eccv20}         & & -     & -     & 54.60 & 50.40 \\
                                 & LVS~\cite{lvs_cvrp21}                   & & -     & -     & 58.20 & 52.50 \\
                                 & EZ-VSL~\cite{ezvsl_eccv22}              & & 19.86 & 30.96 & 62.24 & 54.74 \\
                                 & FNAC~\cite{fnac_cvpr23}                 & & \textbf{35.27} & \underline{38.00} & \underline{84.33} & \underline{63.26} \\
                                 & \cellcolor[gray]{0.9}Ours               & \cellcolor[gray]{0.9}& \cellcolor[gray]{0.9}\underline{35.11} & \cellcolor[gray]{0.9}\textbf{38.66} & \cellcolor[gray]{0.9}\textbf{85.20} & \cellcolor[gray]{0.9}\textbf{65.26} \\ \cline{2-7} 
                                 & EZ-VSL~\cite{ezvsl_eccv22}              & \checkmark & 37.61 & 39.21 & 81.93 & 62.58 \\
                                 & FNAC~\cite{fnac_cvpr23}                 & \checkmark & \textbf{40.97} & \underline{40.38} & 84.73 & 64.34 \\
                                 & \cellcolor[gray]{0.9}Ours               & \cellcolor[gray]{0.9}\checkmark & \cellcolor[gray]{0.9}\underline{39.16} & \cellcolor[gray]{0.9}\textbf{40.51} & \cellcolor[gray]{0.9}\textbf{87.60} & \cellcolor[gray]{0.9}\textbf{66.50} \\ \hline
    \multirow{12}{*}{\makecell{Flickr\\SoundNet\\144k}} & Attention10k~\cite{attention_cvpr18} & & -     & -     & 66.00 & 55.80 \\
                                 & DMC~\cite{dmc_cvpr19}                   & & -     & -     & 67.10 & 56.80 \\
                                 & LVS~\cite{lvs_cvrp21}                   & & -     & -     & 69.90 & 57.30 \\
                                 & HardPos~\cite{hardpos_icassp22}         & & -     & -     & 75.20 & 59.70 \\
                                 & EZ-VSL~\cite{ezvsl_eccv22}              & & 30.27 & 35.92 & 72.69 & 58.70 \\
                                 & FNAC~\cite{fnac_cvpr23}                 & & \underline{33.93} & \underline{37.29} & 78.71 & 59.33 \\
                                 & Alignment~\cite{alignment_iccv23}       & & -     & -     & \underline{85.20} & \underline{62.20} \\
                                 & \cellcolor[gray]{0.9}Ours               & \cellcolor[gray]{0.9}& \cellcolor[gray]{0.9}\textbf{37.01} & \cellcolor[gray]{0.9}\textbf{39.42} & \cellcolor[gray]{0.9}\textbf{86.00} & \cellcolor[gray]{0.9}\textbf{65.16} \\ \cline{2-7} 
                                 & EZ-VSL~\cite{ezvsl_eccv22}              & \checkmark & \underline{41.01} & 40.23 & 83.13 & 63.06 \\
                                 & FNAC~\cite{fnac_cvpr23}                 & \checkmark & \textbf{41.10} & \underline{40.44} & 83.93 & 63.06 \\
                                 & Alignment~\cite{alignment_iccv23}       & \checkmark & -     & -     & \underline{84.00} & \underline{64.16} \\
                                 & \cellcolor[gray]{0.9}Ours               & \cellcolor[gray]{0.9}\checkmark & \cellcolor[gray]{0.9}40.15 & \cellcolor[gray]{0.9}\textbf{40.75} & \cellcolor[gray]{0.9}\textbf{89.20} & \cellcolor[gray]{0.9}\textbf{64.50} \\ 
                                 \bottomrule
    \end{tabular}
    }
    \end{minipage}%
    \hfill
    \begin{minipage}{0.48\linewidth}
        \centering
        \resizebox{\columnwidth}{!}{%
        \begin{tabular}{cc|c|cccc}
        \toprule
        \multirow{2}{*}{\textbf{Trainset}}    & \multirow{2}{*}{\textbf{Method}} & \multirow{2}{*}{OGL} & \multicolumn{2}{c|}{\textbf{VGG-SS}}     & \multicolumn{2}{c}{\textbf{Flickr-SoudNet}} \\
                                     & & & cIoU & \multicolumn{1}{c|}{AUC} & cIoU         & AUC         \\ \hline\hline
        \multirow{7}{*}{\makecell{VGGSound\\10k}}  & LVS~\cite{lvs_cvrp21}     & & -     & -     & 61.80 & 53.60 \\
                                     & EZ-VSL~\cite{ezvsl_eccv22}              & & 25.84 & 33.68 & 63.85 & 54.44 \\
                                     & FNAC~\cite{fnac_cvpr23}                 & & \underline{37.29} & \underline{38.99} & \underline{85.74} & \underline{63.66} \\
                                     & \cellcolor[gray]{0.9}Ours               & \cellcolor[gray]{0.9}& \cellcolor[gray]{0.9}\textbf{37.84} & \cellcolor[gray]{0.9}\textbf{39.82} & \cellcolor[gray]{0.9}\textbf{86.00} & \cellcolor[gray]{0.9}\textbf{64.90} \\ \cline{2-7} 
                                     & EZ-VSL~\cite{ezvsl_eccv22}              & \checkmark & 38.71 & 39.80 & 78.71 & 61.53 \\
                                     & FNAC~\cite{fnac_cvpr23}                 & \checkmark & \underline{40.69} & \underline{40.42} & \underline{82.13} & \underline{63.64} \\
                                     & \cellcolor[gray]{0.9}Ours               & \cellcolor[gray]{0.9}\checkmark & \cellcolor[gray]{0.9}\textbf{41.02} & \cellcolor[gray]{0.9}\textbf{41.29} & \cellcolor[gray]{0.9}\textbf{87.60} & \cellcolor[gray]{0.9}\textbf{65.42} \\ \hline
        \multirow{16}{*}{\makecell{VGGSound\\144k}} & Attention10k~\cite{attention_cvpr18} & & 18.50 & 30.20 & -     & -     \\
                                     & DMC~\cite{dmc_cvpr19}                   & & 29.10 & 34.80 & -     & -     \\
                                     & AVObject~\cite{avobject_eccv20}         & & 29.70 & 35.70 & -     & -     \\
                                     & LVS~\cite{lvs_cvrp21}                   & & 34.40 & 38.20 & 73.50 & 59.00 \\
                                     & HardPos~\cite{hardpos_icassp22}         & & 34.60 & 38.00 & 76.80 & 59.20 \\
                                     & EZ-VSL~\cite{ezvsl_eccv22}              & & 34.38 & 37.70 & 79.51 & 61.17 \\
                                     & SLAVC~\cite{slavc_neurips22}            & & 37.22 & -     & 83.20 & -     \\
                                     & FNAC~\cite{fnac_cvpr23}                 & & 39.50 & 39.66 & \textbf{84.73} & \underline{63.76} \\
                                     & Alignment~\cite{alignment_iccv23}       & & 39.94 & 40.02 & 79.60 & 63.44 \\
                                     & NoPrior~\cite{noprior_cvpr24}           & & \textbf{41.4} & \underline{41.2} & - & - \\
                                     & \cellcolor[gray]{0.9}Ours               & \cellcolor[gray]{0.9}& \cellcolor[gray]{0.9}\underline{40.62} & \cellcolor[gray]{0.9}\textbf{41.66} & \cellcolor[gray]{0.9}\underline{83.60} & \cellcolor[gray]{0.9}\textbf{64.68} \\ \cline{2-7} 
                                     & EZ-VSL~\cite{ezvsl_eccv22}              & \checkmark & 38.85 & 39.54 & 83.94 & 63.60 \\
                                     & SLAVC~\cite{slavc_neurips22}            & \checkmark & 39.67 & -     & \textbf{86.40} & -     \\
                                     & FNAC~\cite{fnac_cvpr23}                 & \checkmark & 41.85 & 40.80 & \underline{85.14} & 64.30 \\
                                     & Alignment~\cite{alignment_iccv23}       & \checkmark & \textbf{42.64} & \underline{41.48} & 82.40 & 64.60 \\
                                     & \cellcolor[gray]{0.9}Ours               & \cellcolor[gray]{0.9}\checkmark & \cellcolor[gray]{0.9}\underline{42.56} & \cellcolor[gray]{0.9}\textbf{42.46} & \cellcolor[gray]{0.9}84.80 & \cellcolor[gray]{0.9}\textbf{64.82} \\ \bottomrule
        \end{tabular}
        }
    \end{minipage}
    \caption{\textbf{Quantitative results of the model trained on Flickr-SoundNet-10k, 144k, and VGGSound-10k, 144k datasets.} Note that the results of previous research are obtained from ~\cite{fnac_cvpr23, alignment_iccv23, noprior_cvpr24}. The results of our method represent the performance after applying Image-Query based Refinement (IQR). The result with OGL denotes the performance after applying object-guided localization ~\cite{ezvsl_eccv22}.}
    \label{table:main_flickr_vggsound}
\end{table*}
\renewcommand{\arraystretch}{1.0}

For the image target slot $\mathbf{p}_i^v$, which is derived from $v_i$, we define its set of $k$-reciprocal nearest neighbors $R_k^v(i)$ as 
\begin{equation}
    R_k^v(i) = \{(v_j, a_j)~|~(j \in N_k^v(i)) \cap (i \in N_k^v(j))\},
\label{eq:knn}
\end{equation}
where $N_k^v(i)$ is the $k$-nearest neighbors of $\mathbf{p}_i^v$, computed using cosine similarity. Likewise, the set of $k$-reciprocal nearest neighbors between audio target slots , i.e., $R_k^a(i)$, can be derived. The predicted false negative set $R_k(i)$ is defined as
\begin{equation}
    R_k(i) = R_k^v(i) \cap R_k^a(i).
\label{eq:rknn}
\end{equation}
Since the samples in $R_k(i)$ have both similar image and audio target slots with $(v_i, a_i)$, they are likely to belong to false negatives. We exclude these predicted false negatives from the contrastive learning in Eq.~\eqref{eq:cotr}. 

\subsection{Inference} \label{method:infer}
For inference, $\textbf{ca}^{a,v}$, which represents the attention between the audio target slot and the input image features, is utilized. Specifically, the attention is interpolated to the original image size, and pixels whose score exceeds the hyperparameter $\theta$ are selected to generate the localization result.

Furthermore, unlike the object-guided localization~\cite{ezvsl_eccv22}, we propose a novel localization refinement strategy, dubbed \textit{Image-Query based Refinement (IQR)}, to refine the localization result without relying on external prior knowledge.
IQR estimates the regions that are likely to contain the target by using the attention between the image target slot and the input image features.
It can be used to identify prior regions in the image that favor the target, regardless of whether these regions are actually the source of the sound.
The refined result derived by IQR, denoted as $R_{\text{IQR}}$, is given as
\begin{equation}
    R_\text{IQR} = \alpha \cdot \textbf{ca}^{a,v} + (1 - \alpha) \cdot \textbf{ia}^{v,v},
\label{eq:refine}
\end{equation}
where $\alpha \in (0, 1)$ %
balances %
the sound source localization result and potential sound source regions.

\section{Experiment}
\begin{table}[!t]
\centering
\scalebox{0.9}{
\begin{tabular}{c|c|cc|cc}
\toprule
\multirow{2}{*}{\textbf{Method}}    & \multirow{2}{*}{OGL} & \multicolumn{2}{c|}{Heard 110} & \multicolumn{2}{c}{Unheard 110} \\
& & cIoU & AUC & cIoU & AUC \\
\hline\hline
LVS~\cite{lvs_cvrp21}                & & 28.90 & 36.20 & 26.30 & 34.70 \\
EZ-VSL~\cite{ezvsl_eccv22}           & & 31.86 & 36.19 & 32.66 & 36.72 \\
SLAVC~\cite{slavc_neurips22}         & & 35.84 & -     & 36.50 & -     \\
Alignment~\cite{alignment_iccv23}   & & \underline{38.31} & \underline{39.05} & \underline{39.11} & \underline{39.80} \\
\cellcolor[gray]{0.9}Ours           &\cellcolor[gray]{0.9} & \cellcolor[gray]{0.9}\textbf{38.46} & \cellcolor[gray]{0.9}\textbf{40.15} & \cellcolor[gray]{0.9}\textbf{40.35} & \cellcolor[gray]{0.9}\textbf{41.25} \\
\cline{1-6}
EZ-VSL~\cite{ezvsl_eccv22}           & \checkmark & 37.25 & 38.97 & 39.57 & 39.60 \\
SLAVC~\cite{slavc_neurips22}         & \checkmark & 38.22 & -     & 38.87 & -     \\
FNAC~\cite{fnac_cvpr23}             & \checkmark & 39.54 & 39.83 & 42.91 & 41.17 \\
Alignment~\cite{alignment_iccv23}   & \checkmark & \textbf{41.85} & \underline{40.93} & \underline{42.94} & \underline{41.54} \\
\cellcolor[gray]{0.9}Ours           & \cellcolor[gray]{0.9}\checkmark & \cellcolor[gray]{0.9}\underline{41.50} & \cellcolor[gray]{0.9}\textbf{41.60} & \cellcolor[gray]{0.9}\textbf{43.11} & \cellcolor[gray]{0.9}\textbf{42.04} \\
\bottomrule
\end{tabular}
}
\caption{\textbf{Quantitative results on Heard 110 and Unheard 110.} Note that the results of previous research are obtained from ~\cite{fnac_cvpr23, alignment_iccv23}. The results of our method represent the performance after applying Image-Query based Refinement (IQR). The result with OGL denotes the performance after applying object-guided localization.}
\label{table:heard_unheard}
\end{table}

\begin{table}[!t]
\centering
\scalebox{0.95}{
\begin{tabular}{cc|cc}
\toprule
    Testset & Method & mIoU & F-score \\ \hline\hline
    \multirow{7}{*}{S4}        & LVS~\cite{lvs_cvrp21}                 & 26.9  & 33.6  \\
                               & EZ-VSL~\cite{ezvsl_eccv22}            & 27.6  & 34.2 \\
                               & SLAVC~\cite{slavc_neurips22}          & 28.1  & 34.6     \\
                               & FNAC~\cite{fnac_cvpr23}               & 27.15 & 31.4 \\
                               & Alignment~\cite{alignment_iccv23}     & \underline{29.6} & \underline{35.9}       \\
                               & Alignment$^\dag$                      & 29.3  & 35.6  \\
                               &\cellcolor[gray]{0.9}Ours                                  &\cellcolor[gray]{0.9}\textbf{31.33} &\cellcolor[gray]{0.9}\textbf{44.30} \\
\bottomrule
\end{tabular}}
\caption{\textbf{Quantitative results on AVSBench.} Note that the results of previous research are obtained from ~\cite{fnac_cvpr23, alignment_iccv23}. $\dag$ denotes that the self-supervised pretrained encoder is utilized for their work.}
\label{table:s4}
\end{table}

\begin{table}[!t]
\resizebox{\columnwidth}{!}{%
\begin{tabular}{c|cccccc}
\toprule
\multirow{2}{*}{Method} & \multicolumn{3}{c|}{Audio $\rightarrow$ Image} & \multicolumn{3}{c}{Image $\rightarrow$ Audio} \\
                        & R@1      & R@5 & \multicolumn{1}{c|}{R@10}  & R@1        & R@5   & R@10   \\ \hline\hline
LVS~\cite{lvs_cvrp21}                   & 3.87              & 12.35             & 20.73             & 4.90              & 14.29             & 21.37       \\
EZ-VSL~\cite{ezvsl_eccv22}                  & 5.01              & 15.73             & 24.81             & 14.20             & 33.51             & 45.18       \\
SSL-TIE~\cite{ssltie_acmmm22}                 & 10.29             & 30.68             & 43.76             & 12.76             & 29.58             & 39.72       \\
SLAVC~\cite{slavc_neurips22}                   & 4.77              & 13.08             & 19.10             & 6.12              & 21.16             & 32.12       \\
Alignment~\cite{alignment_iccv23} & \underline{16.47} & 36.99             & 49.00             & \underline{20.09} & \underline{42.38} & \underline{53.66} \\
Alignment$^\dag$        & 14.31             & \underline{37.81} & \underline{49.17} & 18.00             & 38.39             & 49.02             \\
\rowcolor[gray]{0.9} Ours                    & \textbf{30.61}    & \textbf{57.87}    & \textbf{69.37}    & \textbf{31.70}    & \textbf{58.43}    & \textbf{69.81}    \\
\bottomrule
\end{tabular}
}
\caption{\textbf{Cross-modal retrieval on the VGG-SS.} All of the models are trained on VGGSound-144K. Note that the results of previous research are obtained from ~\cite{alignment_iccv23}. $\dag$ denotes that the self-supervised pretrained encoder is utilized for their work.}
\label{table:retrieval}
\end{table}

\begin{figure*}[t]
    \centering
    \includegraphics[width=0.93\textwidth]{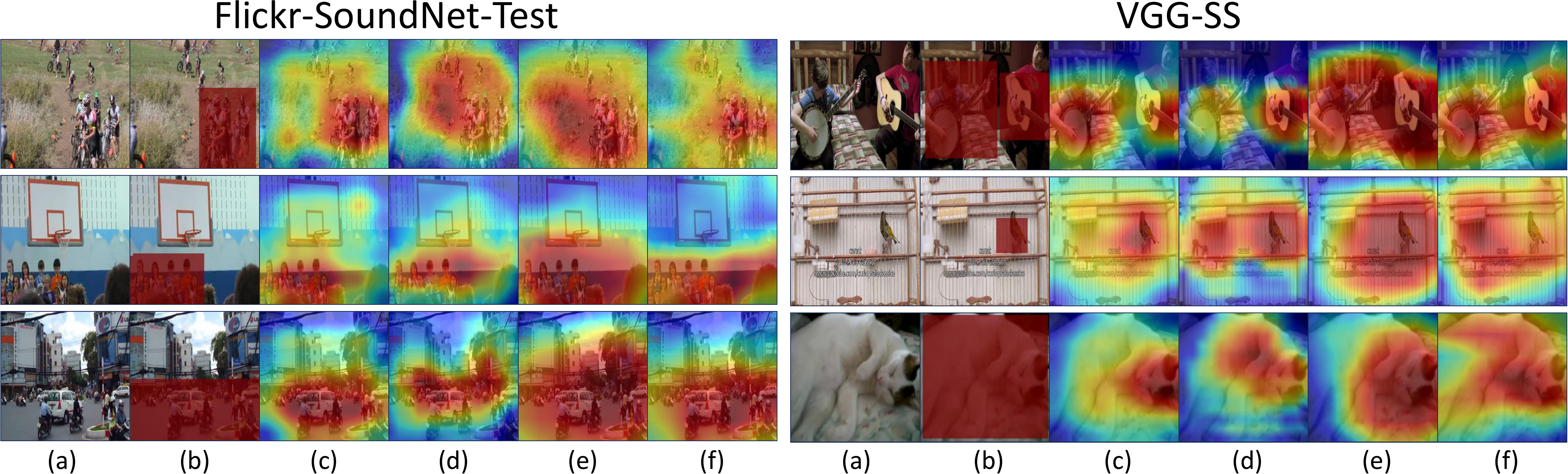}
    \vspace{-2mm}
    \caption{\textbf{Sound localization results on Flickr-SoundNet-Test~\cite{soundnet_neurips16} and VGG-SS~\cite{vggsound_icassp20}.} (a) Input image. (b) Ground-Truth. (c) Ours. (d) Alignment~\cite{alignment_iccv23}. (e) FNAC~\cite{fnac_cvpr23}. (f) EZ-VSL~\cite{ezvsl_eccv22}. The qualitative results are obtained by the model trained on Flickr-144k and the model trained on VGGSound-144k, respectively. Note that all visualizations are obtained without refinement.}
    \label{fig:main_qual}
\end{figure*}

\begin{table}[!t]
\centering
\scalebox{0.82}{
\begin{tabular}{c|ccc|cc}
\toprule
\multirow{2}{*}{Trainset} & \multirow{2}{*}{$\mathcal{L}_{\text{match}}$} & \multirow{2}{*}{$\mathcal{L}_{\text{div}}$}  & \multirow{2}{*}{FN Mitigate} & \multicolumn{2}{c}{VGG-SS}  \\
                               &           &            &            & cIoU  & AUC   \\ \hline\hline
\multirow{8}{*}{\makecell{VGGSound\\144k}} &            &            &            & 14.95 & 27.02 \\
                                           & \checkmark &            &            & 36.23 & 39.13 \\
                                           &            & \checkmark &            & 18.69 & 30.03 \\
                                           &            &            & \checkmark & 15.03 & 28.18 \\
                                           & \checkmark & \checkmark &            & 40.56 & 41.35 \\
                                           & \checkmark &            & \checkmark & 36.14 & 39.31 \\
                                           &            & \checkmark & \checkmark & 20.37 & 30.81 \\
                                           & \checkmark & \checkmark & \checkmark & \textbf{40.71} & \textbf{41.62} \\   
\bottomrule
\end{tabular}
}
\caption{\textbf{Detailed ablation study for losses.}}
\label{table:ablation}
\end{table}

\begin{table}[!t]
\centering
\resizebox{\columnwidth}{!}{%
\begin{tabular}{c|c|cccc}
\toprule
\multirow{2}{*}{Trainset}      & \multirow{2}{*}{\makecell{Shared\\Initial Slots}} & \multicolumn{2}{c|}{VGG-SS}       & \multicolumn{2}{c}{Flickr} \\
                               &            & cIoU  & \multicolumn{1}{c|}{AUC} & cIoU         & AUC         \\ \hline\hline
\multirow{2}{*}{Flickr-144k}   &            & 34.80 & 38.72 & 84.00 & 61.70 \\
                               & \checkmark & \textbf{36.64} & \textbf{39.24} & \textbf{86.80} & \textbf{65.44} \\ \hline
\multirow{2}{*}{VGGSound-144k} &            & 33.11 & 37.92 & 81.20 & 64.36 \\
                               & \checkmark & \textbf{40.71} & \textbf{41.62} & \textbf{84.40} & \textbf{64.50} \\
\bottomrule                               
\end{tabular}
}
\caption{\textbf{Ablation study on shared initial slots.}}
\label{table:slotshare}
\end{table}

\begin{figure}[t]
    \centering
    \includegraphics[width=0.93\columnwidth]{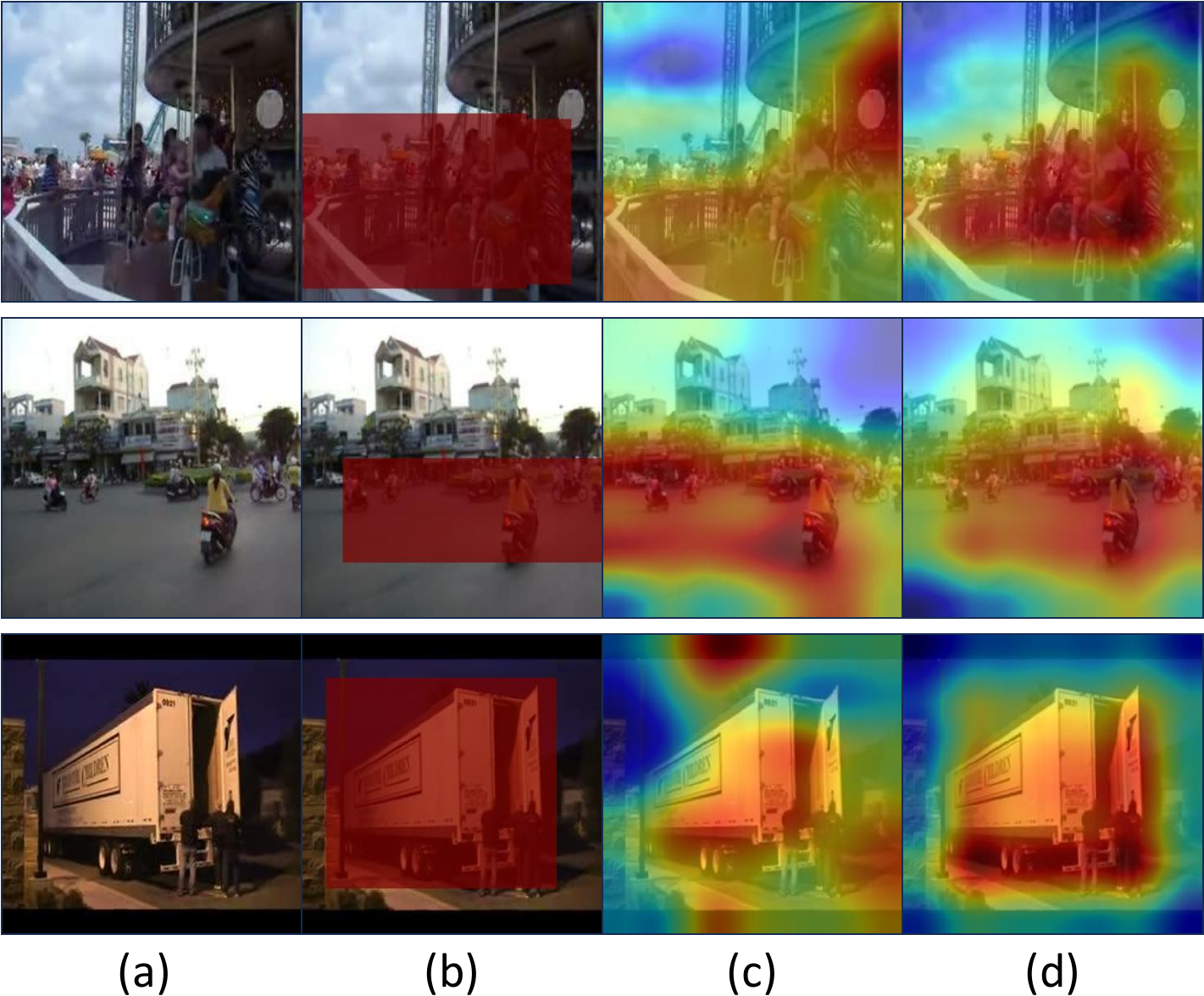}
    \caption{\textbf{Qualitative results to show the impact of $\mathcal{L}_{\text{match}}$ on Flickr-SoundNet.} (a) Input image. (b) Ground-Truth. (c) Results without cross-modal attention matching. (d) Results with cross-modal attention matching.}
    \label{fig:lsim_qual}
\end{figure}

\subsection{Experiment Setup} \label{exp:details}
\textbf{Datasets.} Flickr-SoundNet~\cite{soundnet_neurips16} and VGG-Sound~\cite{vggsound_icassp20} are used to train. Flickr-SoundNet contains about 2 million unconstrained videos from Flickr, while VGG-Sound has around 200k video clips.
For training, we use subsets of 10k and 144k from each dataset, respectively.
For the evaluation, Flickr-SoundNet-Test~\cite{soundnet_neurips16}, VGG-SS~\cite{vggsound_icassp20}, and AVSBench~\cite{avsbench_eccv22} are used following the previous research~\cite{lvs_cvrp21,ezvsl_eccv22,fnac_cvpr23,alignment_iccv23}. Flickr-SoundNet-Test has 250 audio-visual pairs with manually labeled bounding boxes. VGG-SS has about 5k audio-visual pairs over 220 categories. Since VGG-SS provides its category ground truth, it is also used to evaluate cross-modal retrieval performance.
AVSBench has about 5k videos with pixel-wise binary segmentation maps.
Following the previous study~\cite{lvs_cvrp21}, the middle frame of the video clip is used for the image. For the audio, the middle 5 seconds are extracted and converted into log spectrograms with 257 frequency bins at a sampling rate of 16kHz.

\noindent \textbf{Network Architecture.} ResNet-18~\cite{resnet} is adopted for both the image encoder and the audio encoders. The image encoder is fine-tuned from an ImageNet~\cite{Imagenet} pretrained model, while the audio encoder is trained from scratch. Both encoders include an additional 1$\times$1 convolutional layer for projection, and the audio encoder also includes a MaxPooling layer for pooling along the frequency axis. 

\noindent \textbf{Hyperparameter Settings.} We use the AdamW optimizer~\cite{adamw} with a learning rate of 5e-5 and a weight decay of 1e-2. The mini-batch size during training is set to 256, and the size of image and audio feature size $h$, $w$, $t$ and $c$ are set to 7, 7, 16, and 512 respectively. $\tau$ in Eq.~(7) is set to 0.03. $\lambda_1$, $\lambda_2$, and $\lambda_3$ in Eq.~(11) are set to 100.0, 0.1, and 0.1, respectively. $\alpha$ in Eq.~(14) is set to 0.6. The value $k$ in Eq.~\eqref{eq:rknn} is set to 20. The number of iterations $N$ for slot attention mechanism is set to 5.

\noindent \textbf{Evaluation metric.} We measure the average precision at a Consensus Intersection over Union (cIoU) threshold of 0.5 to evaluate localization accuracy. Additionally, the AUC score is used to provide an overall performance measure across different thresholds.

\subsection{Quantitative Results} \label{exp:quan}
\textbf{Flickr-SoundNet and VGGSound.}
In this section, we compare our method with previous self-supervised SSL approaches. We trained our model on subsets of Flickr-SoundNet and VGGSound with dataset sizes of 10k and 144k, and evaluated the performance on Flickr-SoundNet-Test and VGG-SS. 
As shown in Table~\ref{table:main_flickr_vggsound}, our model achieves the state-of-the-art performance in most settings. Specifically, our method yields AUC improvements across all evaluated settings.
Moreover, even without using Object-Guided Localization (OGL), our model performs competitively or better compared to previous methods, demonstrating the robustness of our architecture in focusing on the sound source without additional refinement steps.

\begin{figure*}[t]
    \centering
    \includegraphics[width=0.9\textwidth]{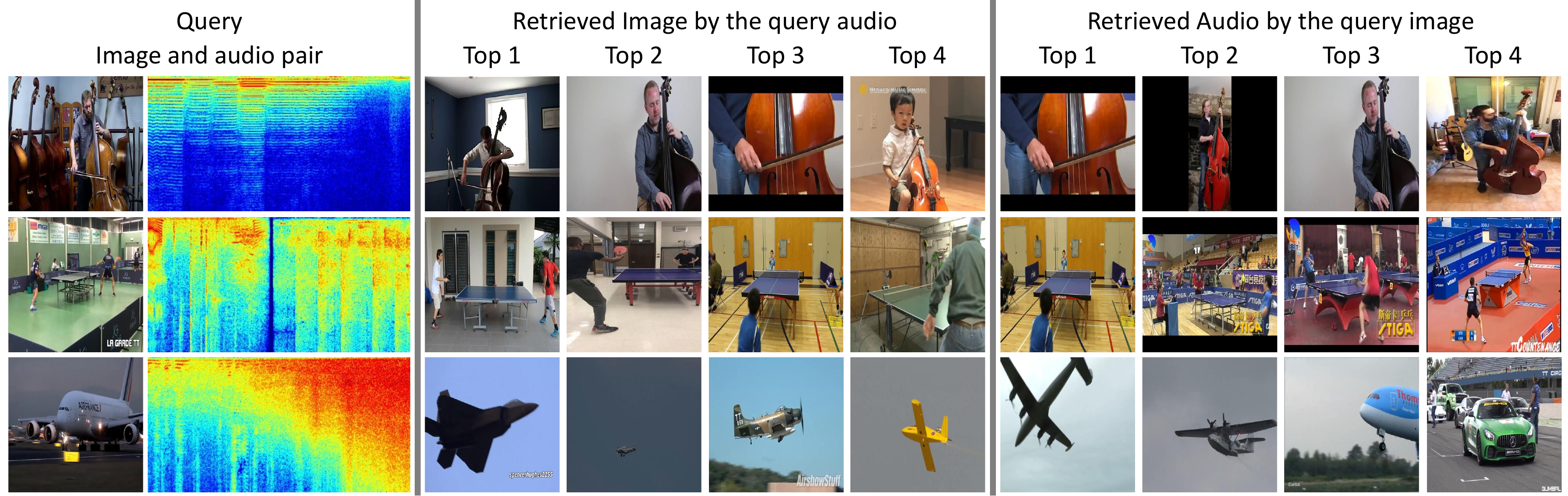}
    \caption{\textbf{Qualitative results of cross-modal retrieval on VGG-SS.} }
    \vspace{-4mm}
    \label{retrieval_qual}
\end{figure*}

\noindent \textbf{Heard 110 and Unheard 110.} To assess the generalization ability of SSL, we evaluate the performance with Heard 110 and Unheard 110.
Heard 110 and Unheard 110 are subsets of the VGG-SS, each containing 110 non-overlapping categories. During training, the model is trained only to the category set of one group. The subset comprising categories seen during training is called Heard 110, while the subset with unseen categories is Unheard 110. As shown in Table~\ref{table:heard_unheard}, our model achieves the state-of-the-art performance on Unheard 110 both without additional refinement and with OGL applied. These results demonstrate that our model has superior generalization ability compared to previous models.

\noindent \textbf{AVSBench.}  
To more precisely evaluate the performance of the localization mask, we measured the performance on AVSBench, which provides segmentation masks. Our model was trained on VGGSound-144k and evaluated in a zero-shot manner on AVSBench. As shown in Table~\ref{table:s4}, we achieved the state-of-the-art results, specifically with improvements of 2.03\%p in mIoU and 8.7\%p in F-score.

\noindent \textbf{Cross-modal Retrieval.} As discussed in the previous study~\cite{alignment_iccv23}, the benchmark evaluation metrics, cIoU and AUC, do not guarantee better audio-visual semantic understanding.
Therefore, we also evaluate cross-modal retrieval performance on VGG-SS.
We extract the image and audio target slot representations and compute the cosine similarity between them.
The similarities are ranked in descending order, and we calculate Recall@K performance using the category labels from the top-K retrieved audio or images.
In Table~\ref{table:retrieval}, our model demonstrates a significant performance improvement over the previous state of the art in cross-modal retrieval. Specifically, our model improves Recall@1 by 16.3\%p in audio-to-image retrieval and 13.7\%p in image-to-audio retrieval. This result indicates that our model does not merely detect salient objects likely to produce sound within the image, but rather effectively locates the target specified by the audio input.

\subsection{Ablation Study} \label{exp:abl}
\textbf{Impact of Losses and False Negative Mitigating in Sec.~\ref{method:migitate}.}
In Table~\ref{table:ablation}, we conduct an ablation study to verify the effectiveness of each loss function and the strategy outlined in Sec.~\ref{method:migitate}.
The overall results show that the designed losses and the false negative mitigating strategy each contribute to improve the performance.
Notably, the experiment with $\mathcal{L}_{\text{match}}$ demonstrates a drastic increase in performance, highlighting the importance of cross-modality attention matching.
Additionally, the ablation study on $\mathcal{L}_{\text{div}}$ shows that explicitly separating the slots enables each slot to perform its designated role more effectively.

\noindent \textbf{Impact of Shared Initial Slots.}
We conduct an ablation study to examine the impact of shared initial slots by using different initial slots for each modality.
As shown in Table~\ref{table:slotshare}, sharing initial slots resulted in better performance, indicating that this design choice facilitates the ability to jointly capture sound source-related information across different modalities.

\subsection{Qualitative Results} \label{exp:qual}
\noindent \textbf{Quality of SSL results.} Fig.~\ref{fig:main_qual} visualizes the qualitative results of our model and previous studies on Flickr-SoundNet~\cite{soundnet_neurips16} and VGG-SS~\cite{vggsound_icassp20} test sets.
These results demonstrate that our model has the ability to effectively localize sound source objects.
Notably, it produces compact localization results even for challenging samples that other models struggle with, such as small birds, a gun, and cars surrounded by other objects in street scenes.

\noindent \textbf{Qualitative Analysis of $\mathcal{L}_{\text{match}}$.}
Fig.~\ref{fig:lsim_qual} shows the qualitative result that shows the effect of $\mathcal{L}_{\text{match}}$ on Flickr-SoundNet-Test.
The attention maps is generated between the audio query of the audio target slot and the image keys derived from image features. 
Without $\mathcal{L}_{\text{match}}$, the alignment between the two is insufficient for compact localization, whereas with $\mathcal{L}_{\text{match}}$, the alignment improves, resulting in more accurate localization of the sound source.

\noindent \textbf{Cross-modal Retrieval.} Fig.~\ref{retrieval_qual} shows the qualitative results of our model for cross-modal retrieval.
Consistent with the impressive performance shown in Table~\ref{table:retrieval}, our model demonstrates a strong ability to retrieve semantically similar samples across modalities that correspond to the sound source target.

\section{Conclusion}

This paper presented a novel self-supervised sound source localization method leveraging joint slot attention to align image and audio features.
Our approach decomposes these features into target and off-target slots, using only target slots for contrastive learning to capture sound-related features.
In addition, we introduced cross-modal attention matching and slot divergence loss which significantly improves localization accuracy and and we demonstrated the effectiveness of our model through comprehensive analysis.
For future work, we aim to extend our framework to multi-source sound source localization.

\noindent\textbf{Acknowledgement.} This work was supported by the NRF grant (RS-2021-NR059830--40\%) and the IITP grants
(RS-2024-00509258--40\%,
RS-2021-II212068--10\%,
RS-2019-II191906--10\%) funded by Ministry of Science and ICT, Korea.

\clearpage
{
    \small
    \bibliographystyle{ieeenat_fullname}
    \bibliography{egbib}
}

\clearpage
\appendix
\section*{Appendix}
\renewcommand{\thesection}{\Alph{section}}
\setcounter{section}{0}

\section{Model Size} \label{supple:model}
Unlike previous studies, we use an additional module for joint slot attention, including auxiliary decoders.
The joint slot attention module consists of 6.84M parameters and the image decoder and audio decoder each have 1.05M parameters.
Compared to EZ-VSL~\cite{ezvsl_eccv22}, which uses only two ResNet-18~\cite{resnet}, our approach requires 39\% more parameters during training, and 30\% more parameters during inference because the decoders are not used for inference.
\begin{table}[!ht]
    \centering
    \resizebox{0.65\columnwidth}{!}{
    \begin{tabular}{c|c}
        \toprule
        Method & \# of parameters (M) \\ \hline \hline
        EZ-VSL~\cite{ezvsl_eccv22} & 22.87 \\
        Ours & 31.82 \\
        $\text{Ours}^{\dag}$ & 29.71 \\
        \bottomrule
    \end{tabular}
    }
    \caption{\textbf{Model size.} $\dag$ denotes the number of used parameters during the inference.}
    \label{table:size}
\end{table}

\section{Experiments on Multi-Source Dataset} \label{supple:multi-source}
To more precisely evaluate the performance of the multi-source dataset, we trained our model on VGGSound-144k and evaluated in a zero-shot manner on AVSBench MS3.
As shown in Table~\ref{table:ms3}, our method showed strong performance on the multi-source dataset and effectively captured distinct objects when two target slots were used.
However, the impact of increasing target slots was limited since the training dataset, VGGSound-144k, is a single-source dataset.

\begin{table}[!t]
    \centering
    \resizebox{0.75\columnwidth}{!}{
    \begin{tabular}{cc|cc}
\toprule
Testset              & Method    & mIoU  & F-Score \\ \hline\hline
\multirow{6}{*}{MS3} & LVS~\cite{lvs_cvrp21}       & 18.54 & 17.4    \\
                     & EZ-VSL~\cite{ezvsl_eccv22}    & 21.36 & 21.6    \\
                     & FNAC~\cite{fnac_cvpr23}      & 21.98 & 22.5    \\
                     & SLAVC~\cite{slavc_neurips22}     & 24.37 & 25.56    \\
                     & \cellcolor[gray]{0.9}Ours$_{(1,1)}$       & \cellcolor[gray]{0.9}\underline{24.45} & \cellcolor[gray]{0.9}\textbf{36.87} \\
                     & \cellcolor[gray]{0.9}Ours$_{(2,1)}$ & \cellcolor[gray]{0.9}\textbf{24.73} & \cellcolor[gray]{0.9}\underline{36.56}          \\
\bottomrule
\end{tabular}
}
    \caption{\textbf{Results on MS3.} 
Subscripts indicate the numbers of target and off-target slots, respectively. }
    \label{table:ms3}
\end{table}

\section{Additional Ablation Studies} \label{supple:ablation}
In this section, we conduct ablation studies on the slot attention iterations $N$, predicted false negatives $k$, masking ratio, the number of each slot. All experiments are trained on VGGSound-144k~\cite{vggsound_icassp20} and tested on VGG-SS~\cite{vggsound_icassp20}.

\noindent \textbf{Slot Attention Iteration.}
We analyze the effect of $N$ by varying slot attention iterations.
Table~\ref{table:combined_iter} shows that 5 iterations achieve the best performance for both sound source localization and cross-modal retrieval. Similar to the findings in the original slot attention~\cite{slotattn_neurips20}, performance declines when the number of iterations is too low or too high.

\begin{table}[!t]
    \centering
    \resizebox{\columnwidth}{!}{%
    \begin{tabular}{c|cc|cccccc}
        \toprule
        \multirow{2}{*}{$N$} & \multirow{2}{*}{cIoU} & \multirow{2}{*}{AUC} & \multicolumn{3}{c|}{Audio $\rightarrow$ Image} & \multicolumn{3}{c}{Image $\rightarrow$ Audio} \\
                            & & & R@1 & R@5 & \multicolumn{1}{c|}{R@10} & R@1 & R@5 & R@10 \\
        \hline \hline
        1  & 16.36 & 27.81 & 1.12  & 5.78  & 9.89  & 3.47 & 11.65 & 17.82 \\
        3  & 39.59 & 40.78 & 16.17 & 36.27 & 47.91 & 19.45 & 41.55 & 53.88 \\
        \rowcolor[gray]{0.9}5 & \textbf{40.71} & \textbf{41.62} & \textbf{30.61} & \textbf{57.87} & \textbf{69.37} & \textbf{31.70} & \textbf{58.43} & \textbf{69.81} \\
        7  & 39.74 & 41.21 & 23.03 & 46.88 & 58.14 & 24.97 & 49.17 & 61.25 \\
        10 & 34.24 & 38.60 & 7.12 & 19.58 & 28.44 & 7.43 & 24.37 & 35.69 \\
        \bottomrule
    \end{tabular}%
    }
    \caption{\textbf{Ablation studies of the number of iterations $N$.} The gray row indicates the settings used in the main paper.}
    \label{table:combined_iter}
\end{table}

\noindent \textbf{Predicted False Negatives.}
We measure the impact of $k$-reciprocal nearest neighbors on false negative mitigation by varying $k$. Table~\ref{table:combined_false} shows that an appropriate $k$ improves performance by excluding false negatives. However, a large $k$ may also exclude true negatives, reducing the number of samples available for contrastive learning and degrading performance. We set $k=20$ for the best results in sound source localization.

\begin{table}[!t]
    \centering
    \resizebox{\columnwidth}{!}{%
    \begin{tabular}{c|cc|cccccc}
        \toprule
        \multirow{2}{*}{$k$} & \multirow{2}{*}{cIoU} & \multirow{2}{*}{AUC} & \multicolumn{3}{c|}{Audio $\rightarrow$ Image} & \multicolumn{3}{c}{Image $\rightarrow$ Audio} \\
                            & & & R@1 & R@5 & \multicolumn{1}{c|}{R@10} & R@1 & R@5 & R@10 \\
        \hline \hline
        10 & 40.29 & 41.16 & \textbf{33.08} & \textbf{60.78} & 71.42 & 33.00 & \textbf{60.06} & \textbf{71.73} \\
        \rowcolor[gray]{0.9}20 & \textbf{40.71} & \textbf{41.62} & 30.61 & 57.87 & 69.37 & 31.70 & 58.43 & 69.81 \\
        30 & 40.15 & 41.14 & 31.87 & 60.06 & \textbf{72.04} & \textbf{33.29} & 59.78 & 71.23 \\
        50 & 40.44 & 41.56 & 28.13 & 55.58 & 67.29 & 29.47 & 56.65 & 68.77 \\
        \bottomrule
    \end{tabular}%
    }
    \caption{\textbf{Ablation studies of $k$-reciprocal nearest neighbors.} The gray row indicates the settings used in the main paper.}
    \label{table:combined_false}
\end{table}

\noindent \textbf{Masking Ratio.} 
During training, we randomly replace 10\% of the input features with learnable mask tokens to prevent overfitting and improve performance.
We investigate the impact of this approach by varying the masking ratio.
Table~\ref{table:combined_maskratio} shows that using a small ratio of learnable masks yields better performance compared to not using masks at all.
However, excessively replacing input features leads to a loss of information, resulting in performance degradation.

\begin{table}[!t]
    \centering
    \resizebox{\columnwidth}{!}{%
    \begin{tabular}{c|cc|cccccc}
        \toprule
        \multirow{2}{*}{Ratio} & \multirow{2}{*}{cIoU} & \multirow{2}{*}{AUC} & \multicolumn{3}{c|}{Audio $\rightarrow$ Image} & \multicolumn{3}{c}{Image $\rightarrow$ Audio} \\
                            & & & R@1 & R@5 & \multicolumn{1}{c|}{R@10} & R@1 & R@5 & R@10 \\
        \hline \hline
        0.0                     & 37.46 & 40.26 & 8.90  & 24.72 & 34.45 & 10.08 & 29.35 & 42.52 \\
        0.05                    & 40.07 & 40.96 & 17.39 & 37.53 & 48.64 & 20.24 & 44.65 & 56.34 \\
        \rowcolor[gray]{0.9}0.1 & \textbf{40.71} & \textbf{41.62} & \textbf{30.61} & \textbf{57.87} & \textbf{69.37} & \textbf{31.70} & \textbf{58.43} & \textbf{69.81} \\
        0.2                     & 33.50 & 38.34 & 9.25  & 23.23 & 32.92 & 9.52  & 27.82 & 39.47 \\
        0.3                     & 31.72 & 37.48 & 9.13 & 22.55 & 31.72 & 9.13 & 27.05 & 39.20 \\
        0.5                     & 16.56 & 29.22 & 12.54 & 30.46 & 40.29 & 15.08 & 36.51 & 48.31 \\
        \bottomrule
    \end{tabular}%
    }
    \caption{\textbf{Ablation studies of the masking ratio.} The gray row indicates the settings used in the main paper.}
    \label{table:combined_maskratio}
\end{table}

\begin{figure}[H]
    \centering
    \includegraphics[width=\columnwidth]{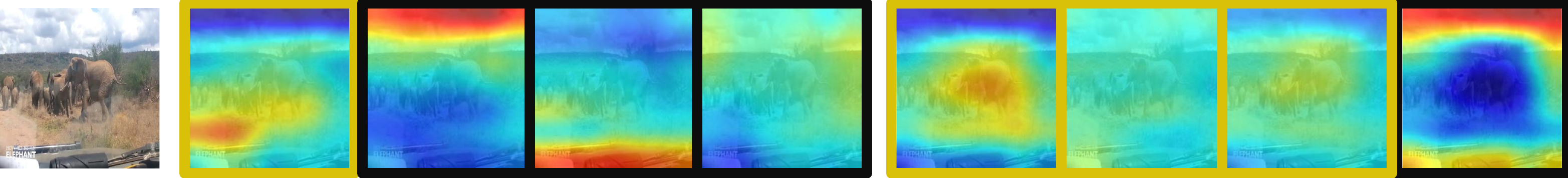}
    \caption{The yellow boundaries indicate target slot attention, while the black boundaries indicate off-target slot attention.
    }
    \label{fig:slot_ablation}
\end{figure}

\noindent \textbf{The Number of Each Slot.}
We analyze the effect of the slots by changing the number of target slot and off-target slot.
Table~\ref{table:num_slots} shows that the performance drops as the number of slots increases. This is mainly because increasing the number of slots too much intensifies their competition in input decomposition. 
For example, with four slots as shown in Fig.~\ref{fig:slot_ablation}, a target area is often fragmented, with no target slot clearly capturing the target, and target or off-target slots tend to intrude into background or target area, respectively.

\section{Failure Cases} \label{supple:failure}
\begin{figure}[!h]
    \centering
    \includegraphics[width=\columnwidth]{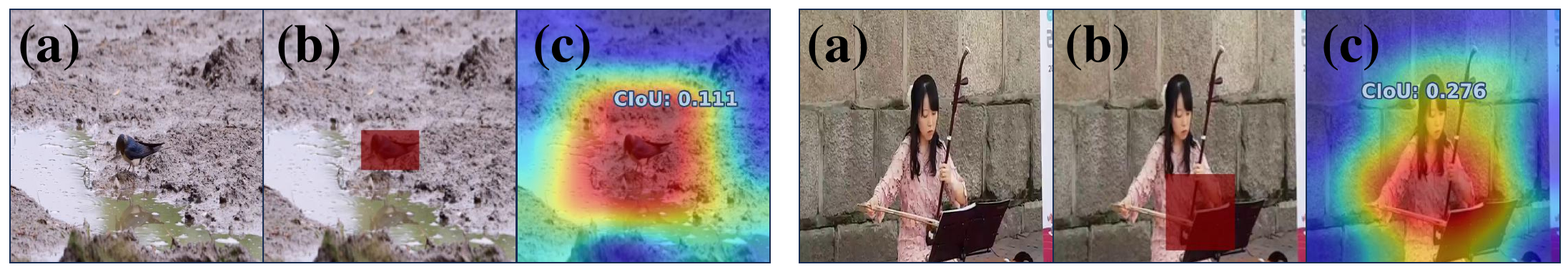}
    \caption{\textbf{Failure cases.} (a) Input (b) Ground-Truth (c) Prediction}
    \label{fig:failure}
\end{figure}

Fig.~\ref{fig:failure} presents failure cases from VGG-SS. Due to the limited resolution of attention, the model sometimes has trouble capturing small objects.
Also, it struggles with strong co-occurrence bias. For example, as instruments and humans frequently co-occur in training data, it sometimes identifies both simultaneously when an instrument sound is given.

\section{More Qualitative Results} \label{supple:qual}
\begin{figure}[H]
    \centering
    \includegraphics[width=\columnwidth]{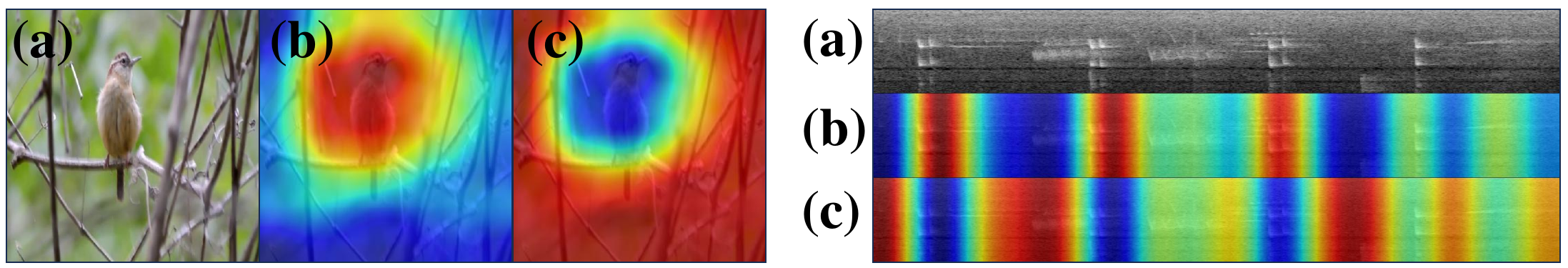}
    \vspace{-7mm}
    \caption{\textbf{Intra-modal attention.} (a) Input image-audio pair (b) Target slot attention (c) Off-target slot attention
    }
    \label{fig:additional_qual}
\end{figure}

\begin{figure*}[t]
    \centering
    \includegraphics[width=\textwidth]{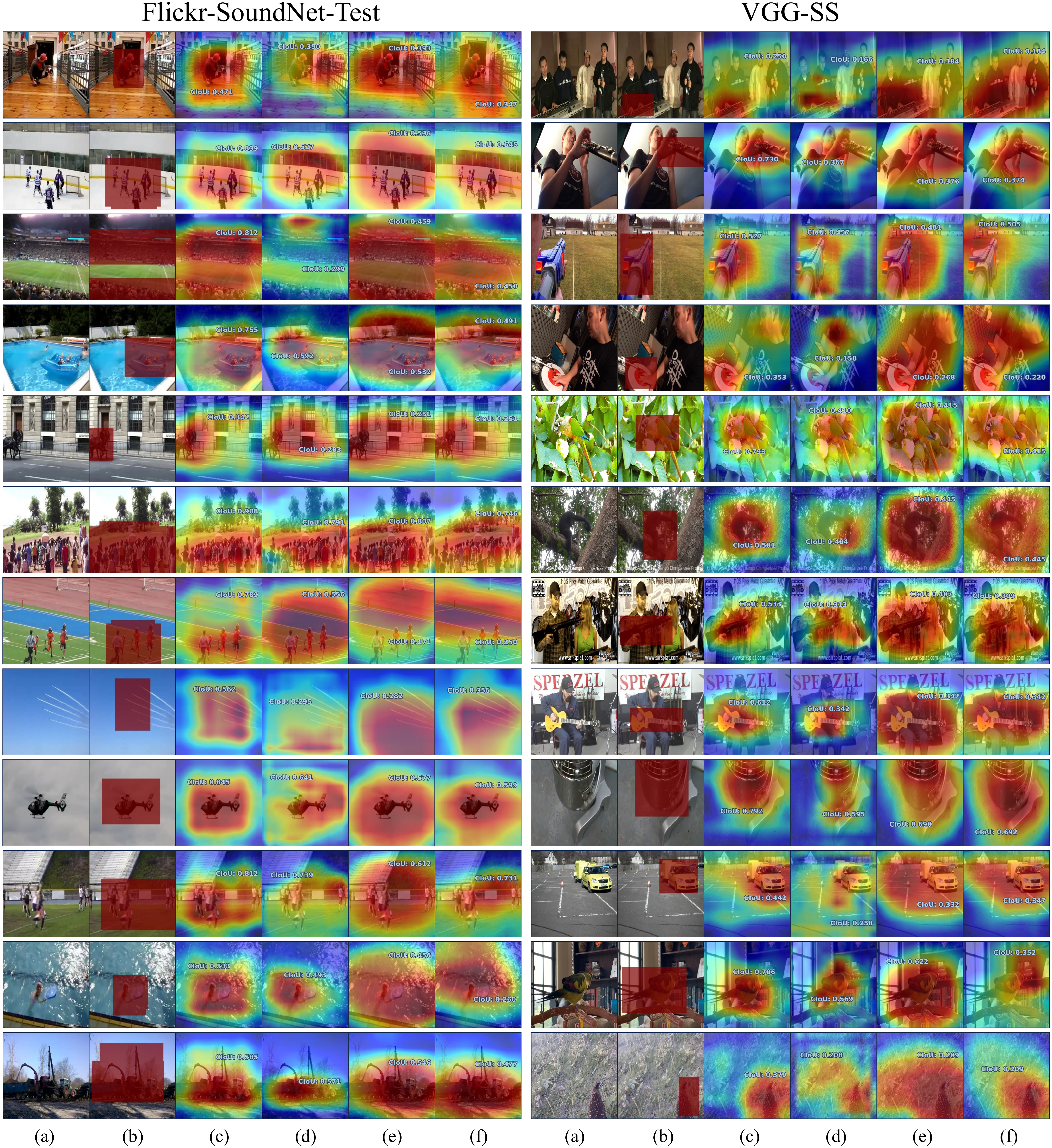}
    \caption{\textbf{Additional sound localization results on Flickr-SoundNet-Test~\cite{soundnet_neurips16} and VGG-SS~\cite{vggsound_icassp20}.} (a) Input image. (b) Ground-Truth. (c) Ours. (d) Alignment~\cite{alignment_iccv23}. (e) FNAC~\cite{fnac_cvpr23}. (f) EZ-VSL~\cite{ezvsl_eccv22}. The qualitative results are obtained by the model trained on Flickr-144k and the model trained on VGGSound-144k, respectively. Note that all visualizations are obtained without refinement.}
    \label{supple:main_qual}
\end{figure*}

\begin{figure*}[t]
    \centering
    \includegraphics[width=\textwidth]{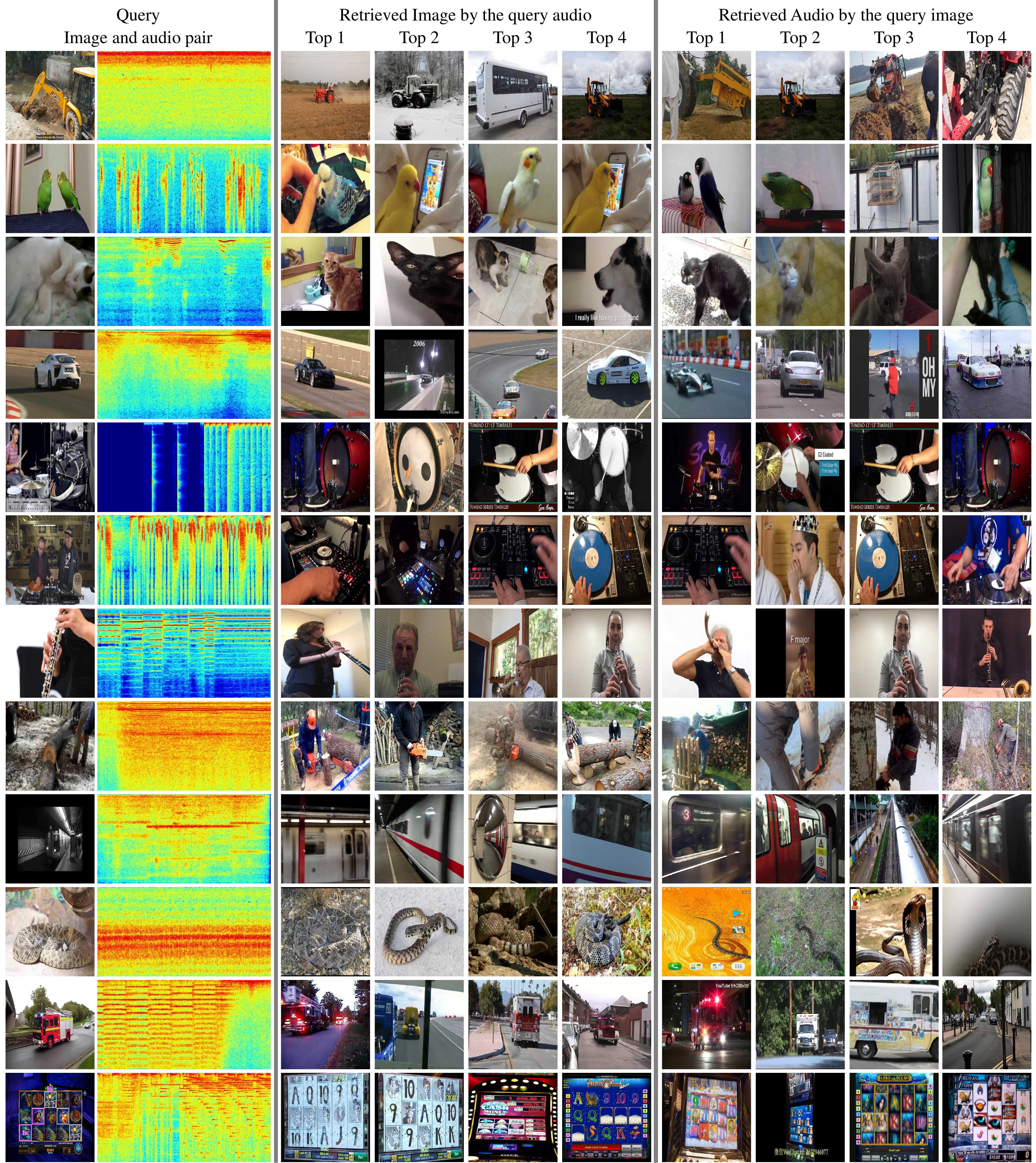}
    \caption{\textbf{Additional Qualitative results of cross-modal retrieval on VGG-SS~\cite{vggsound_icassp20}.} Since the audio spectrograms are difficult to understand, we visualize its corresponding image instead.}
    \label{supple:retrieval}
\end{figure*}

\begin{figure*}[t]
    \centering
    \includegraphics[width=0.9\textwidth]{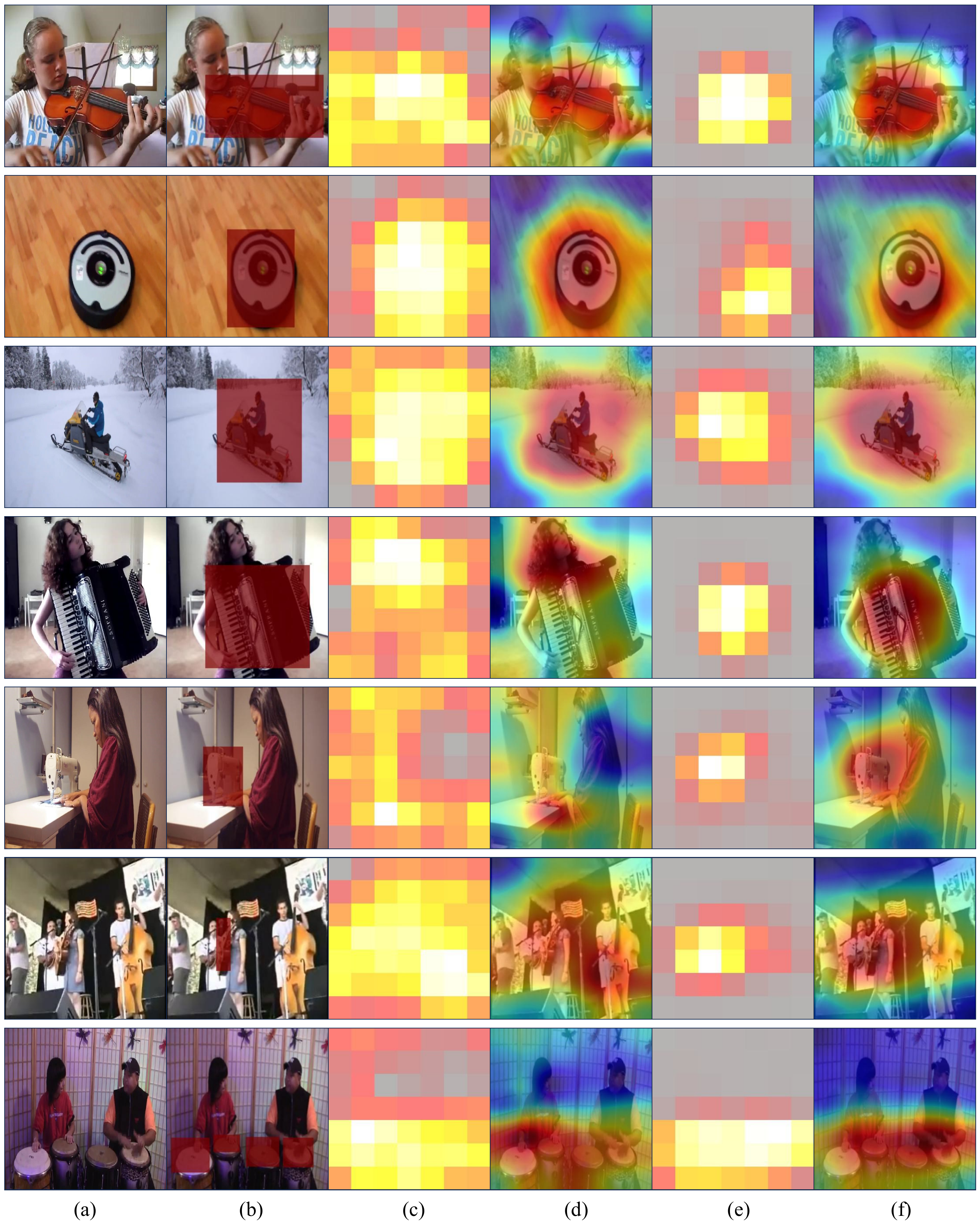}
    \caption{\textbf{Additional qualitative results to show the impact of $\mathcal{L}_{\text{match}}$ on VGG-SS~\cite{vggsound_icassp20}.}
    (a) Input image. (b) Ground-Truth. (c) Attention map of $7\times7$ size without cross-modal attention matching. (d) Attention map of $224\times224$ size without cross-modal attention matching. (e) Attention map of $7\times7$ size with cross-modal attention matching. (f) Attention map of $224\times224$ size with cross-modal attention matching.}
    \label{supple:sim_ablation}
\end{figure*}

\begin{figure*}[t]
    \centering
    \includegraphics[width=\textwidth]{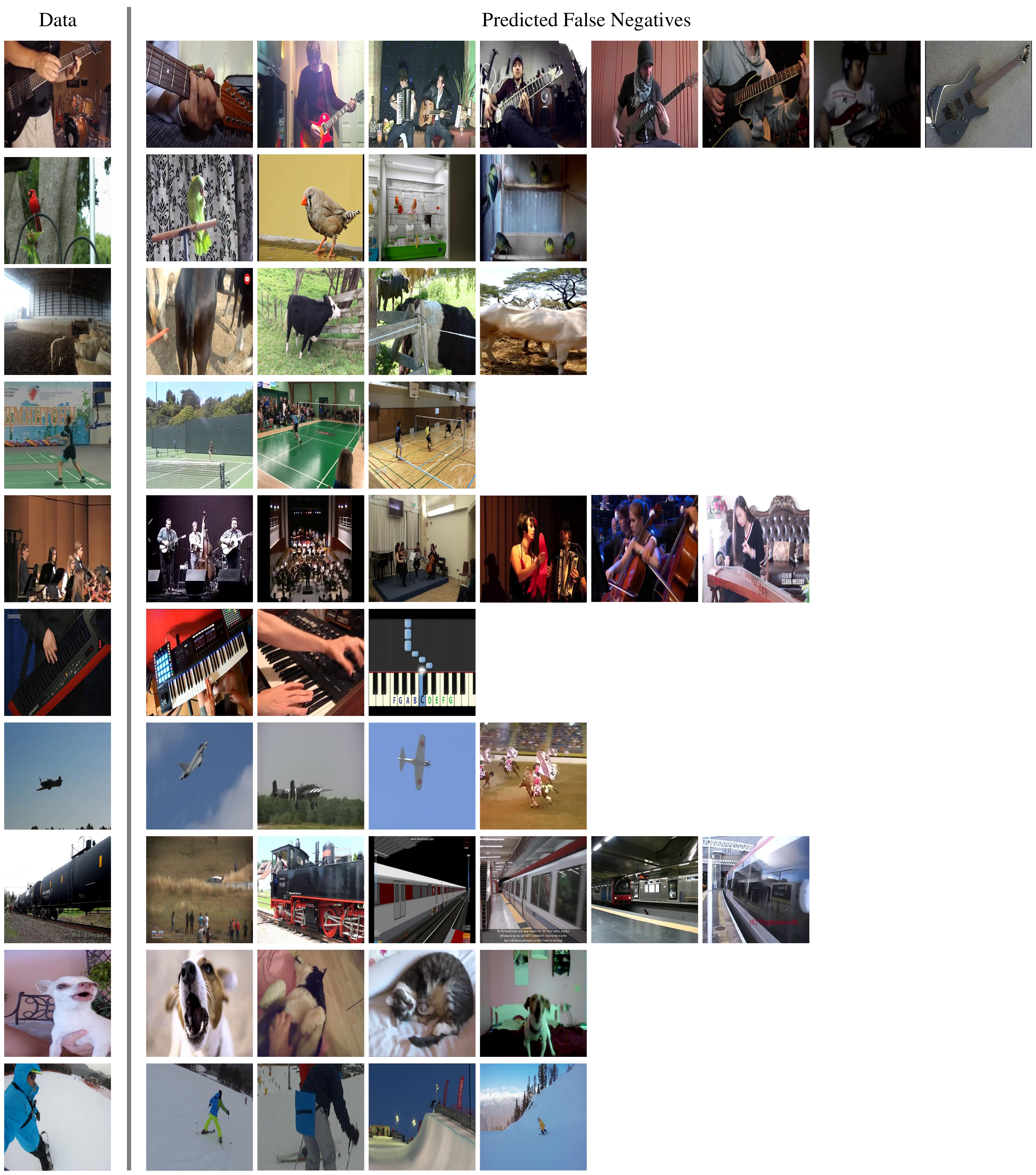}
    \caption{\textbf{Qualitative results of predicted false negative samples of VGGSound~\cite{vggsound_icassp20}.} Samples in the same row are predicted as false negatives by using $k$-reciprocal nearest neighbors within a batch. They have both similar image and audio target slot representations, so they are not used as negative pairs for contrastive learning.}
    \label{supple:pfn}
\end{figure*}

In this section, we present additional qualitative results. Fig.~\ref{fig:additional_qual} shows the intra-modal attention result, showing the target and off-target slots effectively decompose the input by attending to certain regions.
Additionally, in the case of audio, we qualitatively confirmed that the target slot focuses on time intervals related to bird sounds in the spectrogram.
Also, we present additional qualitative results on SoundNet-Flickr-Test~\cite{soundnet_neurips16} and VGG-SS~\cite{vggsound_icassp20} in Fig.~\ref{supple:main_qual}, and additional cross-modal retrieval on VGG-SS in Fig.~\ref{supple:retrieval}.
Furthermore, we visualize the cross-modal attention between the audio target slot and image features with and without $\mathcal{L}_{\text{match}}$ in Fig.~\ref{supple:sim_ablation} to show the effect of $\mathcal{L}_{\text{match}}$. Fig.~\ref{supple:sim_ablation} demonstrates that $\mathcal{L}_{\text{match}}$ encourages the attention map to focus on the sound source.
Finally, Fig.~\ref{supple:pfn} presents some examples of predicted false negatives within a batch.
Fig.~\ref{supple:pfn} demonstrates that $k$-reciprocal nearest neighbors are likely to belong to false negatives.

\vfill\null
\newpage

\begin{table}[H]
    \centering
    \begin{tabular}{cc|cc}
        \toprule
        \multirow{2}{*}{\makecell{Target}} & \multirow{2}{*}{\makecell{Off-Target}} & \multicolumn{2}{c}{VGG-SS} \\
                                           &                                        & cIoU         & AUC         \\ \hline\hline
        \rowcolor[gray]{0.9} 1             & 1                                      & \textbf{40.71} & \textbf{41.62} \\
        1                                  & 2                                      & 38.91 & 40.69 \\
        1                                  & 3                                      & 34.52 & 38.66 \\
        2                                  & 1                                      & 40.36 & 41.17 \\
        3                                  & 1                                      & 37.50 & 40.11 \\
        \bottomrule                               
        \end{tabular}
    \caption{\textbf{Ablation studies of the number of each slots.} The gray row indicates the settings used in the main paper.}
    \label{table:num_slots}
\end{table}

\clearpage

\clearpage

\end{document}